\documentclass[11pt]{article}
\usepackage{coling2020}
\usepackage{times}
\usepackage{latexsym}
\usepackage{amsmath,amsfonts,amssymb,amsthm}
\usepackage{graphicx}
\usepackage{url}
\usepackage{caption}
\usepackage{subfig}
\usepackage{array}
\newcolumntype{?}{!{\vrule width 1pt}}

\colingfinalcopy  

\title{Linguistically inspired morphological inflection with a sequence to sequence model}

\author{Eleni Metheniti* \dag \\
   \\
   \\
   \\
   \\\And
  G{\"u}nter Neumann \ddag \\
   \\
  *CLLE-CNRS, \dag IRIT-CNRS , \ddag DFKI \\
  {\tt eleni.metheniti@univ-tlse2.fr} \\\And
  Josef van Genabith \ddag \\
    \\
    \\
    \\
    \\}

\date{}

\begin{document}
\maketitle
\begin{abstract}
Inflection is an essential part of every human language's morphology, yet little effort has been made to unify linguistic theory and computational methods in recent years. Methods of string manipulation are used to infer inflectional changes; 
our research question is whether a neural network would be capable of learning inflectional morphemes for inflection production in a similar way to a human in early stages of language acquisition.
We are using an inflectional corpus \cite{metheniti2020wikinflection} and a single layer seq2seq model to test this hypothesis, in which the inflectional affixes are learned and predicted as a block and the word stem is modelled as a character sequence to account for infixation. Our character-morpheme-based model creates inflection by predicting the stem character-to-character and the inflectional affixes as character blocks. 
We conducted three experiments on creating an inflected form of a word given the lemma and a set of input and target features, comparing our architecture to a mainstream character-based model with the same hyperparameters, training and test sets. Overall for 17 languages, we noticed small improvements on inflecting known lemmas (+0.68\%) but steadily better performance of our model in predicting inflected forms of unknown words (+3.7\%) and small improvements on predicting in a low-resource scenario (+1.09\%).
\end{abstract}

\section{Introduction}

\label{section:overview}

Inflection is the set of morphological processes that occur in a word, so that the word acquires certain grammatical features which either create syntactic dependencies in a phrase (e.g. gender agreement between nouns and adjectives) or add to the meaning but not change it (e.g. tense in verbs). The parts of speech that are inflected in human languages are commonly nouns, adjectives and verbs, and the set of all possible forms a word can have is its \textit{inflectional paradigm}, where the \textit{lemma} of the word is the canonical form of the word. 

Based on their choices on morphology, languages can be categorized as \textit{isolating} (syntactic categories are expressed with complement words and little/no morphemes, e.g. Chinese), \textit{analytical} (similar to isolating, but slightly morphologically richer, e.g. English) or \textit{synthetic} (use of morphemes to express syntactic properties); the latter can be further distinguished into \textit{fusional} (where an affix may entail more than one meanings, e.g. German), \textit{agglutinative} (where every affix has a unique meaning, 
e.g. Finnish) or polysynthetic (where a word consists of many inflectional affixes, with one or more meanings, e.g. Yupik languages). Other processes may occur inside a word, alongside \textit{affixation} or independently, to express inflectional changes, for example \textit{apophony} in English verbs (r\textbf{i}ng--r\textbf{a}ng) or in German nouns (H\textbf{au}s ``house"--H\textbf{{\"a}u}s-er ``houses"); processes that alter the \textit{stem} create \textit{stem allomorphs}. 

Inflectional processes are learned even from the earliest stages of language acquisition, and native speakers are able to make grammaticality judgements and inflect unknown words. Fully character-based models have been very successful in predicting inflection in a context-free environment, but they are not linguistically motivated.
Our work aims to explore whether a neural architecture would be able to emulate the process of creating inflected word forms in a way similar to humans, given a corpus of morphologically segmented words, their lemmas and their features. We are going to train our models on multiple languages with varying degrees of morphological variety, and for each language we will compare a character-morpheme-based model (learning morphemes as blocks and stems as character sequences) and a character-based model (learning a word as a character sequence), on three experiments with known and unknown lemmas and high and low training scenarios.



\section{Previous Work}
\label{section:previous}

Recent years have shown that interest in morphological analysis, is not waning; there are still challenges to be tackled, areas where morphology can contribute to better results, and tasks to be further explored. 
Computational morphological processing involves many key tasks, such as lemmatization, the process of converting an inflected word to its lemma, and has been approached on a context-free environment  or within its context by either using pre-existing morphological tags \cite{heigold2016neural,putz2018seq2seq}, or inflectional paradigms \cite{bergmanis2019data} or solely the word's neighbouring context \cite{bergmanis2018context,chakrabarty2017context}. However, lemmatization
However, this is a task that does not necessarily capture the morphological choices of the language, since it usually predicts the lemma in absence of the word's inflectional paradigm -- hence the need for context to disambiguate \cite{bergmanis2018context,chakrabarty2017context}. 

On the other hand, generating all of the inflected forms of a lemma can create its inflectional paradigm, as in \newcite{durrett2013supervised}; they collect all the word forms for Wiktionary and map the string transformations from lemma to each word form. With a semi-conditional random field, they applied these rules to unseen cases to generate an `unknown' word's inflectional paradigm. 
Their rule extraction algorithm is able to identify changes in the beginning/end of a string (prefixes/suffixes/circumfixes) and in the middle of the string (infixes or stem allomorphs). However, it is not linguistically aware; 
since it will generate the string transformations based on the given word forms, and the algorithm may dissect the word at its morpheme boundaries but only by coincidence. Also, some transformations they suggest occur in the stem, and in linguistic theory the entire stem is generally treated as a stem allomorph, rather than a string segment that may occur in the middle of the stem. 

\newcite{durrett2013supervised}'s approach to generating inflectional paradigms has been used by \newcite{hulden2014semi}, where they use a small set of concrete inflectional tables to generalise them into inflectional paradigms. First they extract the \textit{longest common substring} (LCS) from the inflection of a lemma, then they create a paradigm out of the remaining string under the assumption that this is the only part relevant to inflection (i.e. contains all the edits needed to create the inflected form), and out of all the paradigms they created, they collapse the ones that are similar to end up with a grammar of substitutions. To generate inflection for new words, they create possible LCSs of the new word and select the paradigms with the most frequent edits. Building on \newcite{hulden2014semi}, \newcite{ahlberg2015paradigm} show that assigning paradigms to new words is more effective when a SVM classifier is used on the lemma substring's features, 
classifying the lemma by its beginning and ending characters. An older approach by \newcite{chrupala2008learning} had already attempted to map differences in substrings by first reversing the strings, in order to find the shortest sequence of insert and delete commands instead of the LCS. \newcite{madsack2018ax}, in one of their experiments for morphological reinflection, drew inspiration from \newcite{ahlberg2015paradigm} alongside by replacing the SVM classifier with 
an algorithm made for computational molecular biology to align and predict.

On the topic of morphological reinflection, recent years have shown an increased interest in the topic through the CoNLL--SIGMORPHON workshop Shared Tasks \cite{cotterell2018conll}, which invite participants to submit systems that can produce target inflected word forms from another inflected form, given a set of source features and target features, in multiple languages and multiple training scenarios (low/medium/high resource). Notable participants of the last four years have been \newcite{kann2016single} with their \textit{sequence-to-sequence encoder-decoder character-model}, a neural approach that takes as input a single character sequence of the source word, the source features and the target features, and outputs the target word. They use \textit{soft attention} \cite{bahdanauCB14} which focuses either copying the input word character-to character for the output, or predicting from the input word if inflection is present, or predicting from all three inputs together. To correct the output, they also use edit trees, if required. This model has since been used as the baseline for the following years' Shared Tasks, because of its overall success for most languages. 
Other papers since then have used different architectures that also build on a sequence-to-sequence model with attention, for example \newcite{acs2018bme} used separate encoders for the source word and the features and two-headed attention.

A neural approach that has been lauded in the Shared Task for its 
with great results on low-resource scenarios was developed by \newcite{aharoni2016improving} (and their follow-up on \newcite{aharoni2016morphological}), who use \textit{hard attention} to generate predictions from only one element of the source sequence instead of the entire sequence, and either predict or copy. In order to assess which characters should be generated or copied in the target sequence, they use a \textit{smart 1-to-0 alignment} to create morphological templates, reminiscent of \newcite{durrett2013supervised}. 
\newcite{makarov2017align} have used their architectures to create their own models of \textit{hard attention with copy mechanism}, which, instead of creating a template, assesses per character whether it should be copied or predicted, and hard \textit{attention over edit action}, which uses edit actions to create the inflected form from the lemma. In \newcite{makarov2018align}, they incorporate the edit distance alignments in the training loss function of their neural model, in order to avoid sub-optimal alignments.

\section{Morphology and Language Acquisition}

Most of the methods presented in Section \ref{section:previous} have proven to perform very well on high-frequency scenarios and on multiple languages with varying morphology; the majority of the systems on the CoNLL--SIGMORPHON 2018 task were able to produce inflections for morphologically-poor and morphological-rich languages with over 90\% accuracy -- with \newcite{makarov2018align} achieving an average of 96\% over all languages \cite{cotterell2018conll}. However, these approaches do not take into account  linguistic knowledge; our research question is, can linguistic insight improve morphological reinflection/induction results, or at least be on par with these purely string manipulation-based methods? In other words, can machine learning learn and produce morphology, in a way similar to human language acquisition? 

Studies in multiple languages have shown that children have morpheme awareness from a very early age; \newcite{SHI1999B11} claim that language acquisition starts already at birth, with infants acquiring functional elements based on frequency and sound properties. During the holophrastic period of language acquisition (9-18 months), infants are able to comprehend monomorphemic words, functional words and word stems, even though they are not able to produce complex speech. \newcite{mintz} found out that 15-month-olds are able to discern the `-ing' suffix in English, and \newcite{MARQUIS201261} claim that 11-month-old children could also distinguish the French past participle suffix `-{\'e}' by inferring its presence comparing the lemma and inflected term, even in non-words. In the two-word stage (18-24 months), toddlers begin to form simple sentences and they can make grammaticality judgements; even though they may make production mistakes such as omitting articles \cite{gerken1990function}, they are not able to process incorrect structures, meaning that they are aware of the underlying morphosyntactic bonds between words, word order, morpheme meaning and morpheme productivity (\newcite{santelmann1998sensitivity} for English, \newcite{oshima-takane_ariyama_kobayashi_katerelos_poulin-dubois_2011} for Japanese, \newcite{hohle} for German toddlers). Regardless of the morphological richness of the language, it has been theorised by experimental research that in these primary stages of language acquisition, humans have morpheme representations in their mental lexica, thus enabling them to understand and produce existing or non-existing words (neologisms, loan words, comprehension of nonce words).

The process of morpheme inference and learning in humans, although based on phonetic rather than written representations, is somewhat similar to the LCS process that an algorithm would follow -- with the exception that it is commonly suggested in linguistic theories that humans learn stem allomorphs and not sequences of character or phoneme substitutions in stem. 
If our model inferred morphemes from strings \textit{a priori}, the process would return string substitutions that are not linguistically motivated, therefore for our research question we need to provide the morpheme boundaries for the model to learn human language morphemes.
The first step to explore our hypothesis would be to find a resource of inflected forms, of significant size, preferably multilingual, with gold-standard morpheme boundaries. We would need the inflected forms to be already segmented into stems and inflectional morphemes, and in addition, we would like to explore as many languages as possible, with different levels of morphological richness, to investigate how inflectional diversity will affect our results; according to psycholinguistic studies, there is a positive correlation between a high number of inflected forms per lemma in a language and the speed of language acquisition in children \cite{xanthos2011role}.


\section{Methodology}

\subsection{Finding the inflectional information}
\label{section:method}
Corpora that include inflectional information are not widely available; only a few corpora have a significant number of entries that are annotated for morphological inflections (e.g. the T{\"u}bingen Treebank of Written German; \newcite{telljohann2004tuba}, Korpus 2000 for Danish; \newcite{asmussen2001korpus}, Corpus Of Serbian Language (CSL) for Serbian; \newcite{kostic2001quantitative}, Stockholm Ume{\aa} Corpus; \newcite{ejerhed2006stockholm}). 

We \cite{metheniti2018wikinflection,metheniti2020wikinflection} have created an inflectional corpus for 138 languages with information pulled from the English Wiktionary; the Wiktionary community has created inflectional templates for many languages in order to automatically generate the inflectional tables for lemmas. In our work \cite{metheniti2018wikinflection}, we use these inflectional templates in a similar way to create the inflectional paradigms of lemmas, where every lemma is categorized by template and language, and the word forms are segmented to stem (allomorph) and prefixes, suffixes and infixes if available. However, we noticed some problems with this corpus; first of all, it does not include some languages with a large number of lemmas (e.g. English, French, Portuguese, Arabic), because Wikiflection uses the Wiktionary \textit{templates} to generate the inflection which are not accessible for all languages. Second, in some Wiktionary templates, some function words are included in the word form as affixes, e.g. `no' in Spanish (the function word for negation), in cases that they affect the word form or affect the morphosyntactic features of the word, thus we also use them as affixes. Also, again due to the syntax of the templates, affixes are merged together in agglutinative languages and not separated as shown in Table \ref{tab:koula}, which renders their processing similar to fusional languages and will not allow us to explore how multiple affixes in a sequence are learned. Function words are also included in some templates, such as auxiliary verbs for perfect tenses and negation words, and are included in Wikiflection as prefixes.

\begin{table}[!htb]
\centering
\begin{tabular}{llll}
\hskip-0.2cm koula & \hskip-0.2cm``to coach" & \hskip-0.2cmkoula-isi$_{+[cond]}$-n$_{+[1st\: sing.]}$ & \hskip-0.2cm``if I coach"\\
\hskip-0.2cm trainier-en & \hskip-0.2cm``to coach" & \hskip-0.2cmtrainier-est$_{+[subj] +[2nd\: sing.]} $ & \hskip-0.2cm``I would coach"
\end{tabular}
\caption{Comparing the morphology of Finnish and Spanish. To create a verb form, Finnish uses a suffix for mood, a suffix for tense (no suffix needed for Present tense) and a suffix for person and number. Spanish on the other hand has one suffix to signify mood, tense, person and number.}
\label{tab:koula}
\end{table}

In order to deal with these issues, and also with the additional problem that the generated Wikinflection feature tags are problematic (since some tags are missing due to HTML parsing problems), we generated the Wikinflection corpus and corrected it, by evaluating using the paradigms and feature tags of UniMorph 3.0 \cite{mccarthy2020unimorph}. UniMorph is an ongoing project of creating a large morphological corpus from Wiktionary in a similar way to Wikinflection, with supervised feature-annotated inflectional paradigms over 150 languages, however, it does not include morpheme boundaries. Our Wikinflection corpus \cite{metheniti2020wikinflection} is the intersection of these two corpora, using the words and morpheme boundaries from Wikinflection and the features from UniMorph, converted to the Universal Dependencies V2 notation \cite{11234/1-2895}. Out of this, we kept languages that had more than 10.000 total types. 


For our experiments, we we are going to test each inflected form individually and not the entire paradigm. First, we randomly held out 100 lemmas of each language, to be used in Experiment 2 (see Section \ref{section:results}) as a test set of `unseen' lemmas and their inflected word forms. Then, we created the rest of the lemma (plus features) and target inflected word (plus features) tuples, and we separated them to the training set and test set for Experiments 1 and 3 with an 80/20 ratio of lemmas. A full list of the languages we used is provided in Table \ref{tab:numbers}, and out of 17 languages, 5 are agglutinative (Estonian (est), Finnish (fin), Hungarian (hun), Georgian (kat), Northern Sami (sme)) and 12 are fusional (Old English (ang), Danish (dan), German (deu), Faroese (fao), Irish (gle), Latin (lat), Latvian (lav), Lithuanian (lit), Macedonian (mkd), Polish (pol), 
Swedish (swe), Classic Syriac (syc)).

\begin{table}[!htb]
\centering
\begin{tabular}{|p{1cm}|p{1.5cm}|p{1.25cm}|p{1.5cm}|p{1.5cm}|p{1cm}|p{1.5cm}|p{1.5cm}|p{1.3cm}|}
\hline
\textbf{ISO 639-3} & \textbf{Unseen Lemmas} & \textbf{Unseen Words} & \textbf{Seen Lemmas} & \textbf{Seen Words} & \textbf{Test Words} & \textbf{Train Words} & \textbf{Total Lemmas} & \textbf{Total Words}  \\ \hline
\textbf{ang} & 100 & 1325 & 2420 & 31147 & 484 & 30663 & 2520 & 32472 \\ \hline
\textbf{dan} & 100 & 200 & 5599 & 11198 & 1119 & 10079 & 5699 & 11398 \\ \hline
\textbf{deu} & 100 & 1168 & 4631 & 52354 & 926 & 51428 & 4731 & 53522 \\ \hline
\textbf{est} & 100 & 4541 & 914 & 44045 & 182 & 43863 & 1014 & 48586 \\ \hline
\textbf{fao} & 100 & 1722 & 3311 & 56338 & 662 & 55676 & 3411 & 58060 \\ \hline
\textbf{fin} & 100 & 3361 & 56487 & 1864166 & 4729 & 1852869 & 56587 & 1867527 \\ \hline
\textbf{gle} & 100 & 3234 & 9722 & 339220 & 1944 & 337276 & 9822 & 342454 \\ \hline
\textbf{hun} & 100 & 285 & 19567 & 69030 & 3913 & 65117 & 19667 & 69315 \\ \hline
\textbf{kat} & 100 & 2064 & 3876 & 79808 & 775 & 79033 & 3976 & 81872 \\ \hline
\textbf{lat} & 100 & 1404 & 13351 & 185985 & 2670 & 183315 & 13451 & 187389 \\ \hline
\textbf{lav} & 100 & 2200 & 3156 & 69339 & 631 & 68708 & 3256 & 71539 \\ \hline
\textbf{lit} & 100 & 1498 & 1176 & 18643 & 235 & 18408 & 1276 & 20141 \\ \hline
\textbf{mkd} & 100 & 875 & 3544 & 32227 & 708 & 31519 & 3644 & 33102 \\ \hline
\textbf{pol} & 100 & 3590 & 3716 & 131028 & 743 & 130285 & 3816 & 134618 \\ \hline
\textbf{sme} & 100 & 2647 & 1644 & 44797 & 328 & 44469 & 1744 & 47444 \\ \hline
\textbf{swe} & 100 & 754 & 1443 & 10859 & 288 & 10571 & 1543 & 11613 \\ \hline
\textbf{syc} & 100 & 2444 & 1545 & 37405 & 309 & 37096 & 1645 & 39849 \\ \hline
\end{tabular}
\caption{Each corpus size, with the size of the train set, the test set, and the number of unseen lemmas and unseen words. The languages are named using the ISO 639-3 language codes, as in \newcite{kirov2016very}
}
\label{tab:numbers}
\end{table}

\subsection{Sequence-to-sequence model}
\label{section:seq}
As discussed in Section \ref{section:previous}, it has been shown that sequence-to-sequence models are very effective neural architectures for morphologically related tasks, whether context is included or not. Sequence-to-sequence models are often encoder-decoder models using a recurrent neural network (RNN, LSTM or GRU units) to encode the source (input) sequence(s) into a single vector (context vector), which is then decoded by a second RNN to generate the output of the model, sequence-by-sequence (words, characters, timestamps etc). 

\begin{figure}[!htb]
    \centering
    \includegraphics[width=0.8\textwidth]{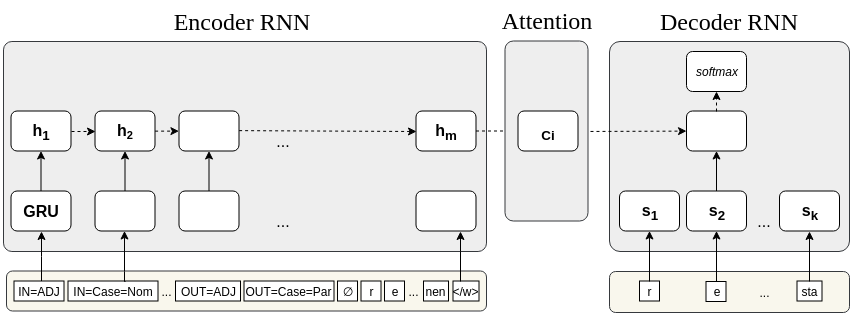}
    \caption{The process of predicting the word `retrogradista', given the input features, the target features and the lemma (full inputs and output can be found at Table \ref{table:inputs}.)}
    \label{fig:model}
\end{figure}

We are using an architecture similar to \newcite{kann2016single} for our model; we encode input as a single string of input features, target features and input word, and expect the target word as output. The input word is the lemma, the input features are the lemma features, the target features are the output's features and the output is the expected word, given the target features. All elements of the input sequence are encoded together. The format of the input and output can be seen in Table \ref{table:inputs}; our novelty lies in training with characters and inflectional morphemes treated as one character, so that the model predicts entire prefixes, suffixes and infixes, and the stem character-by-character to account for allomorphy. To create naive alignments, we have encoded the empty prefix and suffix positions with the use of the \textit{zero morpheme}. For example, the word `retrogradinen' in Finnish is segmented as `retrogradi-nen' (=retrograde), and the inflected form of singular number, partitive case is `retrogradi-sta' (=`with some retrograde').

\begin{table}[!htb]
\centering
\footnotesize
\begin{tabular}{ll}
\hskip-0.2cm Input: & \hskip-0.2cm IN=ADJ IN=Case=Nom IN=Number=Sing \\
 & \hskip-0.2cm OUT=ADJ OUT=Case=Par OUT=Number=Sing \\
 & \hskip-0.2cm $\emptyset$ r e t r o g r a d i nen \textless\textbackslash w\textgreater \\
\hskip-0.2cm Output: & \hskip-0.2cm $\emptyset$ r e t r o g r a d i sta
\end{tabular}
\caption{Sample inputs and output of out current architecture. The input is one string, split by spaces. The `IN' tags refer to the input features (lemma), the `OUT' tags to the target features. The lemma is split by characters in the stem and by morphemes in the affixes, and same applies for the output.}
\label{table:inputs}
\end{table}

Our model is a single-layer encoder-decoder with GRU units with attention; we are using dot-product attention as described in \newcite{luong2015effective}, which gets the decoder hidden state at time \textit{t} and calculates the context vector $\mathbf{h_t}$ which will be transposed and concatenated with the hidden state of the decoder $\mathbf{\bar{h}_s}$, to create the prediction. The attention score is given in Equation \ref{eq:dot}. 

\begin{equation}
    score(\mathbf{h_t}, \mathbf{\bar{h}_s}) = \mathbf{h_t}^\intercal \mathbf{\bar{h}_s}
    \label{eq:dot}
\end{equation}

We use the same hyperparameters for all languages; batch size of 20, hidden size of 100, embedding size of 300, and Adadelta as optimizer. All languages were trained on 20 epochs. An overview of our model's architecture can be seen in Figure \ref{fig:model}.

\section{Results}
\label{section:results}
In order to test our original hypothesis, whether morphemes can be effectively learned by a neural network, we have designed three experiments, to test the quality of predictions in different settings. For each of the 17 languages in Table \ref{tab:numbers} we will train both a character-morpheme-based model and a purely character based model, with the exact same architecture, hyperparameters and training/test sets (segmented accordingly), to determine which one performs best. For example, we will train the Finnish character-morpheme model (fin) as described in Table \ref{table:inputs} in Section \ref{section:seq}, and the character-based model (fin\_char) with inputs and output as seen in Table \ref{table:char2char}. 

\begin{table}[!htb]
\centering
\footnotesize
\begin{tabular}{ll}
\hskip-0.2cm Input: & \hskip-0.2cm IN=ADJ IN=Case=Nom IN=Number=Sing \\
 & \hskip-0.2cm OUT=ADJ OUT=Case=Par OUT=Number=Sing \\
 & \hskip-0.2cm r e t r o g r a d i n e n \textless\textbackslash w\textgreater \\
\hskip-0.2cm Output: & \hskip-0.2cm r e t r o g r a d i s t a
\end{tabular}

\caption{Sample inputs and output for the character-based model.}
\label{table:char2char}
\end{table}

In order to measure the performance and to compare the character-morpheme-based model with the character-based model, we are first going to measure how many predictions were right and how many were wrong, based on the expected output. This measure is effective in showing us how good the models are in predicting words, but we would like to provide a more fine-grained and informative measure of the quality of the predictions, even the ones that were incorrect; a false prediction that is wrong by many characters should be more problematic than a prediction that is wrong by one character. Thus, we also evaluate our results by calculating the Levenshtein distance for every tuple of expected output and prediction, as implemented in the \texttt{python-Levenshtein} python package \cite{levenshtein}, and specifically, the function \texttt{ratio}; this function returns a percentage of similarity between two strings based on the character insertions, deletions and substitutions, and is calculated with the function in Equation \ref{equation:ratio} for words $w_1$, $w_2$ of length $|w_1|$ and $|w_2|$ respectively. It is important to note that this function calculates Levenshtein distance in a way similar to the Longest Common Subsequence (LCS) problem; while an insertion or deletion edit has a cost of 1, a substitution edit has a cost of 2 because, according to LCS, a substitution is composed of a deletion and an insertion. Some examples of the ratio function for the German word `Haus' and possible edits can be seen in Table \ref{tab:ratio}. To get one average number for every language and model, we calculate the similarity ratio for all expected words-predictions, and average it over the size of the test set.

\begin{equation}
\small
    ratio(w_1, w_2) = \frac{(|w_1| + |w_2|) - lev_{a,b}(|a|, |b|)}{|w_1| + |w_2|}
\label{equation:ratio}
\end{equation}

\begin{table}[!htb]
\small
    \centering
    \begin{tabular}{ll}
        $ratio($`Haus', `Hause'$)$ & $0.8888888888888888$ \\
        $ratio($`Haus', `Hau'$)$ & $0.8571428571428571$ \\
        $ratio($`Haus', `Haas'$)$ & $0.75$ \\ 
        $ratio($`Haus', `Haase'$)$ & $0.6666666666666666$ \\ 
        $ratio($`Haus', `Haa'$)$ & $0.5714285714285714$ \\ 
    \end{tabular}
    \caption{The Levenshtein similarity ratio, as computed by Equation \ref{equation:ratio}.}
    \label{tab:ratio}
\end{table}

\subsection{Experiment 1: Seen lemmas, unseen words}

The first experiment we are conducting is training our models on the training set for every language, and predicting words from the test set (see Table \ref{tab:numbers} for the sizes). This test set is a subset of the `seen' words, meaning that the input words are lemmas which are also present in the training set; therefore the model has already `seen' the lemma and knows how to predict a word form X (therefore it probably knows under which inflectional category the lemma falls), and is asked to predict a word form Y. Given the large size of the training sets (significantly larger for most languages, compared to the 10K high resource training scenarios for the CoNLL--SIGMORPHON Shared Task), we expect both the character-morpheme-based models and the character-based models to perform well. The results for this experiment can be found in Table \ref{tab:results}. As expected, both models perform very well for 15 out of 17 languages (over 96\% correct predictions with both models for 10 languages and over 96\% similarity accuracy for 15 languages), with the character-morpheme-based model predicting word forms marginally better for 7 languages, the character-based model being marginally better for 8 languages, and the models tying for 2 languages. The Levenshtein similarity ratio of the predictions shows that the character-morpheme-based model is closer to accurate predictions for 10 languages, but again the difference is marginal and statistically insignificant. The only language that showed statistically significant improvement with the use of the character-morpheme-based model is Classic Syriac, regarding the number of correct predictions (+8.7\%).

\begin{figure}[!htb]

    \centering
    \subfloat[est]{{\includegraphics[width=5.5cm]{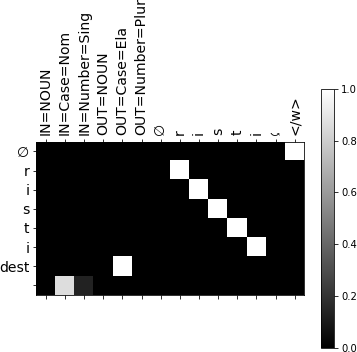} }}%
    \qquad
    \subfloat[est\_char]{{\includegraphics[width=5.5cm]{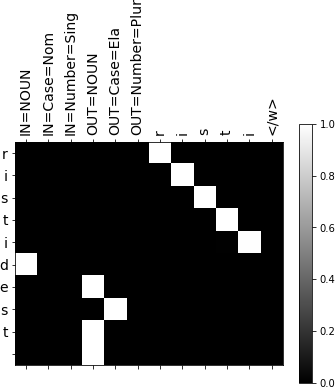} }}%
        
    \caption{The Estonian noun \textit{risti} ``clubs", in elative plural: \textit{risti-de}$_{[+plural]}$\textit{-st}$_{[+elative]}$}%
    \label{fig:test-est}%
\end{figure}

\begin{figure}[!htb]

    \centering
    \subfloat[lit]{{\includegraphics[width=5.5cm]{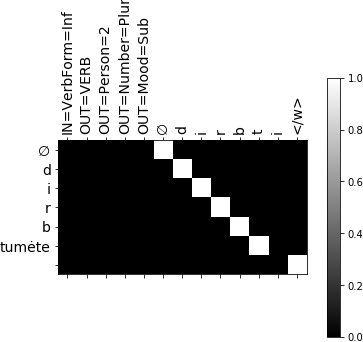}}}%
    \qquad
    \subfloat[lit\_char]{{\includegraphics[width=5.5cm]{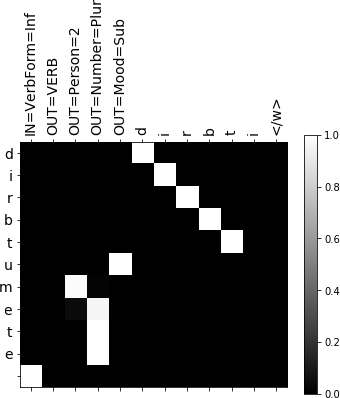} }}%
        
    \caption{The Lithuanian verb \textit{dirb-ti} ``to work", in 2nd person plural subjunctive: \textit{dirb-tum\.{e}te}$_{[+2nd \:plur. \:+subj]}$}
    \label{fig:test-lit}
\end{figure}

For an in-depth look on how the models predict words, we will examine the attention plots of predictions from an agglutinative language, Estonian, and a fusional language, Lithuanian. In Estonian, our model (which had as input the stem character-per-character and the suffixes as one suffix as explained in Section \ref{section:method}) was able to predict the stem based on the lemma's stem and the final suffixes based on the output feature `Case=Ela' (Figure \ref{fig:test-est}a), whereas the character-based model predicted the suffixes' characters from most of the output features. For the character-morpheme-based model, it is an ideal scenario to associate morphemes with certain morphological features, but it is not always the case and not in all languages, as our results show. Examining Lithuanian, for example, in Figure \ref{fig:test-lit}, we notice that the prediction of the suffix \textit{-tum\.{e}te} in our model is aligned with the lemma and not with the output features, but in the character-based-model the attention during the prediction of the characters of the suffix is scattered among the output features.

\subsection{Experiment 2: Unseen lemmas, unseen words}

The second experiment we conducted uses the same trained models and methodology from Experiment 1, to predict words from unseen lemmas. As mentioned in Section \ref{section:method}, we held out 100 lemmas and their corresponding inflected forms from every language before we extracted the training and test sets. These are words that exist in the language, but are `unknown' to the models. Our experiment aims to replicate how native speakers are able to adapt neologisms and non-words to pre-existing inflectional categories of their languages, and create inflectional paradigms that previously did not exist in their language and make grammaticality judgements. 

\begin{figure}[!htb]

    \centering
    \subfloat[hun]{{\includegraphics[width=5.5cm]{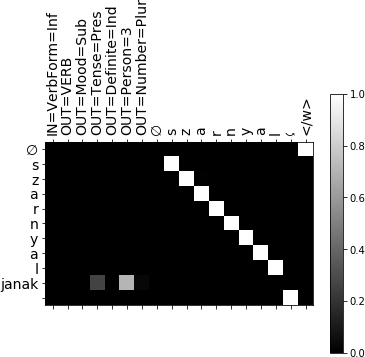} }}%
    \qquad
    \subfloat[hun\_char]{{\includegraphics[width=5.5cm]{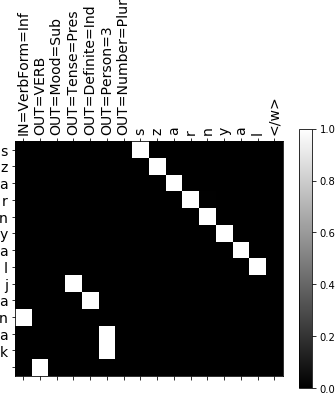} }}%
        
    \caption{The Hungarian verb \textit{sz{\'a}rnyal} ``to soar", in 3rd person plural, present indefinite subjunctive: \textit{sz{\'a}rnyal-j}$_{[+subj.]}$\textit{-anak}$_{[+3rd\: plur.]}$.}
    \label{fig:unseen-hun}%
\end{figure}

\begin{figure}[!htb]

    \centering
    \subfloat[lat]{{\includegraphics[width=5.5cm]{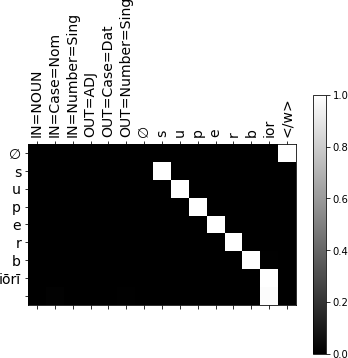}}}%
    \qquad
    \subfloat[lat\_char]{{\includegraphics[width=5.5cm]{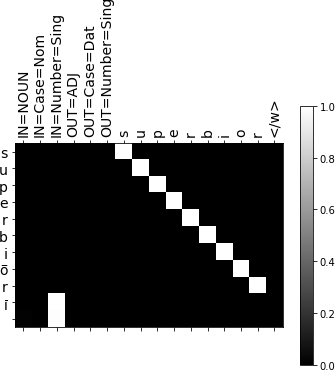} }}%
        
    \caption{The Latin adjective \textit{superbior} ``prouder": \textit{superbior-i}$_{[+2nd \:plur. \:+subj]}$.}
    \label{fig:unseen-lat}
\end{figure}

The results for this experiment can be found in Table \ref{tab:results} as well. As expected, the amount of correct predictions is lower for both models for all languages, compared to Experiment 1, due to the added difficulty of classifying an unseen lemma to a learned inflectional paradigm and predicting its inflected form, however, 4 languages (ang, fao, lit, syc) still manage to have over 96\% correct predictions even in this scenario. Concerning our two models, the character-morpheme-based model outperforms the character-based model in 15 out of the 17 languages, but only statistically significantly in 5 languages. Observing the prediction similarity, we notice that accuracy was marginally better for 16 languages and, interestingly enough, the quality of the predictions was better for the 2 languages (fin and kat) for which the character-based model made more correct predictions, but was lower for 1 language (est) whose character-morpheme-based model made more correct predictions. The attention plots for Hungarian (Figure \ref{fig:unseen-hun}) and Latin (Figure \ref{fig:unseen-lat}) provide similar observations on predictions as their counterparts in Experiment 1. 

\subsection{Experiment 3: ``Poverty of the stimulus"}

Our final experiment draws inspiration both from low-resource scenarios in natural language processing and from the hypothesis that children, even by being exposed to a limited amount of words, are able to make inferences that allow them to learn their native language's grammar. While this theory has been criticised and reviewed over the years, it could be interesting to see how our models perform on the same languages and test sets, given a much smaller training set. 
We used the same 17 languages, but trained with a set of 3000 words instead of the original training sizes. We also prepared a test set of a maximum 500 words for each language and model, with randomly selected words from the test set. These words, because of the small size of the test set, will either be known or unknown lemmas to the model. The results of this experiment can also be found in Table \ref{tab:results}. 

\begin{figure}[!htb]
    \centering
    \subfloat[fin]{{\includegraphics[width=5.5cm]{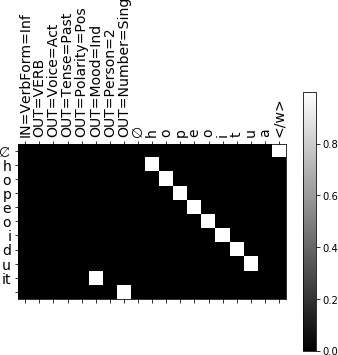} }}%
    \qquad
    \subfloat[fin\_char]{{\includegraphics[width=5.5cm]{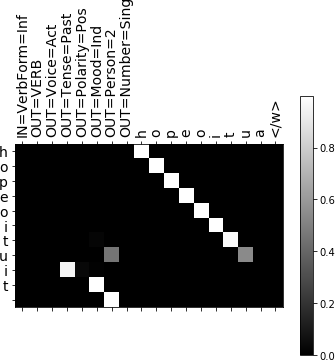} }}%
        
    \caption{The Finnish verb \textit{hopeoitu-a} ``to be silver plated", in 2nd person singular, past positive indicative: \textit{hopeoi\textbf{d}u-i}$_{[+past]}$\textit{-t}$_{[+2nd\: sing.]}$.}
    \label{fig:pov-fin}%
\end{figure}

As expected, we notice a large drop in prediction correctness, for most languages. Some languages still perform adequately well, which would be expected from some less morphologically-rich fusional languages (dan, swe), while some morphologically-rich agglutinative languages perform very well (hun) or relatively well (kat) or very bad (fin). However, as shown in Figure \ref{fig:pov-fin}, our model successfully predicted the stem allomorphy (\textit{t}$\rightarrow$\textit{d}) and correctly focused attention on the stem and not on the features (because this stem allomorph is not dependent to morphological features), even though the model performed badly for the language.

\begin{table*}[]
\tiny
\begin{tabular}{?l?l|p{0.4cm}|p{0.5cm}|l|l?l|p{0.4cm}|p{0.5cm}|l|l?l|p{0.4cm}|p{0.5cm}|l|l?}
\hline
\multicolumn{1}{?c?}{\textbf{}} & \multicolumn{5}{c?}{\textbf{Experiment 1}} & \multicolumn{5}{c?}{\textbf{Experiment 2}} & \multicolumn{5}{c?}{\textbf{Experiment 3}} \\ \hline
\textbf{ISO 639-3} & \textbf{Test} & \textbf{Right} & \textbf{Wrong} & \textbf{Acc} & \textbf{Lev} & \textbf{Test} & \textbf{Right} & \textbf{Wrong} & \textbf{Acc} & \textbf{Lev} & \textbf{Test} & \textbf{Right} & \textbf{Wrong} & \textbf{Acc} & \textbf{Lev} \\ \hline
\textbf{ang} & 484 & \textbf{422} & 62 & \textbf{0.8719} & \textbf{0.9863} & 1325 & \textbf{892} & 433 & \textbf{0.6732\dag} & \textbf{0.9490} & 483 & \textbf{353} & 130 & \textbf{0.7308} & \textbf{0.9623} \\ 
\textbf{ang\_char} & 484 & 414 & 70 & 0.8554 & 0.9838 & 1325 & 797 & 528 & 0.6015 & 0.9236 & 483 & 350 & 133 & 0.7246 & 0.9481 \\ \hline
\textbf{dan} & 1119 & \textbf{1115} & 4 & \textbf{0.9964} & \textbf{0.9996} & 200 & \textbf{198} & 2 & \textbf{0.9900} & \textbf{0.9995} & 500 & 446 & 54 & 0.892 & 0.9904 \\ 
\textbf{dan\_char} & 1119 & 1113 & 6 & 0.9946 & 0.9992 & 200 & 189 & 11 & 0.9450 & 0.9941 & 500 & \textbf{454} & 46 & \textbf{0.908} & \textbf{0.9913} \\ \hline
\textbf{deu} & 926 & 907 & 19 & 0.9795 & 0.9964 & 1168 & \textbf{1047} & 121 & \textbf{0.8964} & \textbf{0.9869} & 500 & \textbf{404} & 96 & \textbf{0.808} & \textbf{0.9642} \\ 
\textbf{deu\_char} & 926 & \textbf{911} & 15 & \textbf{0.9838} & \textbf{0.9969} & 1168 & 1006 & 162 & 0.8613 & 0.9734 & 500 & 385 & 115 & 0.77 & 0.9575 \\ \hline
\textbf{est} & 182 & 173 & 9 & 0.9505 & 0.9911 & 4541 & \textbf{3026} & 1515 & \textbf{0.6664} & 0.9494 & 179 & \textbf{144} & 35 & \textbf{0.8047\dag} & \textbf{0.9691} \\ 
\textbf{est\_char} & 182 & 173 & 9 & 0.9505 & \textbf{0.9915} & 4541 & 2986 & 1555 & 0.6576 & \textbf{0.9540} & 179 & 133 & 46 & 0.7430 & 0.9575 \\ \hline
\textbf{fao} & 585 & \textbf{533} & 52 & \textbf{0.9111} & \textbf{0.9810} & 1660 & \textbf{1215} & 445 & \textbf{0.7319\dag} & \textbf{0.9375} & 500 & \textbf{282} & 218 & 0.564 & 0.9136 \\ 
\textbf{fao\_char} & 585 & 528 & 57 & 0.9026 & 0.9790 & 1693 & 1110 & 583 & 0.6556 & 0.9151 & 446 & 271 & 175 & \textbf{0.6076} & \textbf{0.9169} \\ \hline
\textbf{fin} & 4729 & 4581 & 148 & 0.9687 & \textbf{0.9971} & 3361 & 3250 & 111 & 0.9670 & \textbf{0.9971} & 500 & 85 & 415 & 0.17 & 0.8421 \\ 
\textbf{fin\_char} & 4729 & \textbf{4612} & 117 & \textbf{0.9753} & 0.9972 & 3361 & \textbf{3258} & 103 & \textbf{0.9694} & 0.9966 & 500 & \textbf{291} & 209 & \textbf{0.582\dag} & \textbf{0.9393\dag} \\ \hline
\textbf{gle} & 1944 & 1935 & 9 & 0.9954 & 0.9993 & 3234 & \textbf{3024} & 210 & \textbf{0.9351} & \textbf{0.9901} & 500 & 428 & 72 & 0.856 & 0.9790 \\ 
\textbf{gle\_char} & 1944 & \textbf{1938} & 6 & \textbf{0.9969} & \textbf{0.9994} & 3234 & 3017 & 217 & 0.9329 & 0.9868 & 500 & \textbf{440} & 60 & \textbf{0.88} & \textbf{0.9829} \\ \hline
\textbf{hun} & 3068 & 3027 & 41 & 0.9866 & 0.9986 & 213 & 206 & 7 & 0.9671 & \textbf{0.9964} & 500 & \textbf{455} & 45 & \textbf{0.91} & \textbf{0.9854} \\ 
\textbf{hun\_char} & 3068 & \textbf{3030} & 38 & \textbf{0.9876} & \textbf{0.9989} & 213 & 206 & 7 & 0.9671 & 0.9890 & 500 & 451 & 49 & 0.902 & 0.9789 \\ \hline
\textbf{kat} & 775 & 679 & 96 & 0.8761 & \textbf{0.9756} & 2064 & 1746 & 318 & 0.8459 & \textbf{0.9683} & 500 & \textbf{410} & 90 & \textbf{0.82\dag} & \textbf{0.9626} \\ 
\textbf{kat\_char} & 776 & \textbf{680} & 97 & \textbf{0.8763} & 0.9732 & 2064 & \textbf{1751} & 313 & \textbf{0.8484} & 0.9645 & 500 & 381 & 119 & 0.762 & 0.9457 \\ \hline
\textbf{lat} & 2670 & \textbf{2664} & \textbf{6} & \textbf{0.9978} & \textbf{0.9997} & 1392 & \textbf{1086} & 306 & \textbf{0.7802\dag} & \textbf{0.9736} & 500 & \textbf{365} & 135 & \textbf{0.73\dag} & \textbf{0.9614} \\ 
\textbf{lat\_char} & 2670 & 2634 & 36 & 0.9865 & 0.9976 & 1392 & 927 & 465 & 0.6659 & 0.9500 & 500 & 300 & 200 & 0.6 & 0.9363 \\ \hline
\textbf{lav} & 631 & 629 & 2 & 0.9968 & 0.9993 & 2200 & \textbf{2192} & 8 & \textbf{0.9964} & \textbf{0.9996} & 500 & \textbf{491} & 9 & \textbf{0.982} & \textbf{0.9960} \\ 
\textbf{lav\_char} & 631 & \textbf{631} & 0 & \textbf{1.00} & \textbf{1.00} & 2200 & 2156 & 44 & 0.9800 & 0.9975 & 500 & 489 & 11 & 0.978 & 0.9956 \\ \hline
\textbf{lit} & 235 & 206 & 29 & 0.8766 & \textbf{0.9779} & 1498 & \textbf{1022} & 476 & \textbf{0.6822\dag} & \textbf{0.9360} & 231 & 138 & 93 & 0.5974 & 0.9132 \\ 
\textbf{lit\_char} & 235 & \textbf{207} & 28 & \textbf{0.8809} & 0.9751 & 1498 & 834 & 664 & 0.5567 & 0.9068 & 231 & \textbf{149} & 82 & \textbf{0.6450} & \textbf{0.9237} \\ \hline
\textbf{mkd} & 708 & 412 & 296 & 0.5819 & \textbf{0.9587} & 875 & \textbf{505} & 370 & \textbf{0.5771} & \textbf{0.9508} & 500 & \textbf{274} & 226 & \textbf{0.548} & \textbf{0.9512} \\ 
\textbf{mkd\_char} & 708 & \textbf{416} & 292 & \textbf{0.5876} & 0.9537 & 875 & 499 & 376 & 0.5703 & 0.9446 & 500 & 271 & 229 & 0.542 & 0.9430 \\ \hline
\textbf{pol} & 725 & \textbf{718} & 7 & \textbf{0.9903} & \textbf{0.9992} & 3520 & \textbf{2869} & 651 & \textbf{0.8151} & \textbf{0.9681} & 485 & 359 & 129 & 0.7402 & \textbf{0.9449} \\ 
\textbf{pol\_char} & 725 & 713 & 12 & 0.9834 & 0.9946 & 3520 & 2792 & 728 & 0.7932 & 0.9523 & 485 & \textbf{369} & 116 & \textbf{0.7608} & 0.9432 \\ \hline
\textbf{sme} & 328 & 317 & 11 & 0.9665 & 0.9962 & 2647 & \textbf{1872} & 775 & \textbf{0.7072} & \textbf{0.9630} & 328 & 249 & 81 & 0.7591 & 0.9525 \\ 
\textbf{sme\_char} & 328 & 317 & 11 & 0.9665 & \textbf{0.9970} & 2647 & 1787 & 860 & 0.6751 & 0.9605 & 328 & \textbf{260} & 68 & \textbf{0.7927} & \textbf{0.9696} \\ \hline
\textbf{swe} & 288 & \textbf{286} & 2 & \textbf{0.9931} & \textbf{0.9986} & 754 & \textbf{601} & 153 & \textbf{0.7971} & \textbf{0.9705} & 288 & 260 & 28 & 0.9028 & \textbf{0.9807} \\ 
\textbf{swe\_char} & 288 & 283 & 5 & 0.9826 & 0.9965 & 754 & 579 & 175 & 0.7679 & 0.9472 & 288 & 260 & 28 & 0.9028 & 0.9629 \\ \hline
\textbf{syc} & 309 & \textbf{162} & 147 & \textbf{0.5242\dag} & 0.9029 & 2444 & \textbf{1110} & 1334 & \textbf{0.4542\dag} & \textbf{0.8863} & 309 & \textbf{159} & 150 & \textbf{0.5146\dag} & \textbf{0.9048\dag} \\ 
\textbf{syc\_char} & 309 & 135 & 174 & 0.4369 & 0.8768 & 2444 & 987 & 1457 & 0.4038 & 0.8681 & 309 & 13 & 296 & 0.0421 & 0.6351 \\ \hline
\textbf{Total} &  &  &  & \textbf{0.9096} & \textbf{0.9857} &  &  &  & \textbf{0.7931} & \textbf{0.9660} &  &  &  & \textbf{0.7253} & \textbf{0.9514} \\ 
\textbf{Total char} &  &  &  & 0.9028 & 0.9830 &  &  &  & 0.7560 & 0.9544 &  &  &  & 0.7143 & 0.9369 \\ \hline
\end{tabular}
\caption{Results for all 3 experiments, for all 17 languages. With bold font are marked the best results per language between the two models, and with \dag \:are marked the results that have statistically significant improvement ($p>=0.05$). \textbf{Total} are the average percentages for all languages for the character-morpheme-based models and \textbf{Total char} for the character-based ones.}
\label{tab:results}
\end{table*}

\clearpage
\section{Discussion}

The results for Experiment 1 show that there is no drastic improvement in accuracy or quality of predictions with the use of our character-morpheme-based model. This could be expected, due to the fact that our training sets are adequately large for the character-based model to learn to predict correctly the changes in characters given the morphological features. Still, our model managed to stay on par with the character-based model, even given the additional information that it had to learn (e.g. the suffix `\textit{-s}' in Danish was learned with a different encoding than the character `\textit{s}').  

In Experiment 2, however, we notice a marginal yet overall improvement with our model; this task required the model to be able to classify the lemma in the correct inflectional category and predict the correct morphemes/characters. For some languages (ang, fao, lat, lit, syc) there is a statistically significant improvement over the character-based model; this could be due to the fact that these are morphologically-rich fusional languages, where correct classification of the lemma is crucial to correctly generate the paradigm, therefore learning the entire morpheme in correlation to the morphological features and the stem worked in our favour.

The results of Experiment 3 are interesting, but not necessarily definitive. We expected all morphologically-rich languages (especially agglutinative) to perform much worse, because of the high number of inflectional morphemes that may not be seen in the small training corpus. We have to consider that Finnish and Classic Syriac are outliers, because of the vast differences in accuracy and quality, but for most languages, we see either improvement or marginal underperformance of our model. This finding is quite interesting, and is surely worth further exploration with more languages and different test sets in the future. 

Our experiments were in agreement with our original hypothesis, whether it would be possible and beneficial for a machine-learning architecture to learn morphological knowledge the way a human would. Our goal was not to discredit the string manipulation approaches that have been previously used, rather to explore the intersection between human language acquisition and machine learning. We found out that there is an improvement not just in accuracy, but in the overall quality of predictions, with errors occurring mostly in the prediction of the stem. 
In our future research, we would like to explore whether different neural architectures would eliminate these errors, and what tasks would benefit from the use of our corpus and results.


\newpage

\bibliographystyle{coling}
\bibliography{conll-2019}

\end{document}